\documentclass{article}
\usepackage{spconf,amsmath,epsfig}
\usepackage{tcolorbox}
\let\OLDthebibliography\thebibliography
\renewcommand\thebibliography[1]{
  \OLDthebibliography{#1}
  \setlength{\parskip}{0pt}
  \setlength{\itemsep}{0pt plus 0.3ex}
}
\begin{document}\sloppy

\def\x{{\mathbf x}}
\def\L{{\cal L}}

\title{Temporal Insight Enhancement\protect: Mitigating Temporal Hallucination in Multimodal Large Language Models}

\name{Li Sun$^{1,2}$ \,Liuan Wang$^{1}$ \, Jun Sun$^{1}$ \, Takayuki Okatani$^{2,3}$ }

\address{$^{1}$ Fujitsu R$\&$D Center, Beijing, China \\
    $^{2}$Graduate School of Information Sciences, Tohoku University
    $^{3}$RIKEN Center for AIP
    }

\maketitle

\begin{abstract}
Recent advancements in Multimodal Large Language Models (MLLMs) have significantly enhanced the comprehension of multimedia content, bringing together diverse modalities such as text, images, and videos. However, a critical challenge faced by these models, especially when processing video inputs, is the occurrence of hallucinations – erroneous perceptions or interpretations, particularly at the event level. This study introduces an innovative method to address event-level hallucinations in MLLMs, focusing on specific temporal understanding in video content. Our approach leverages a novel framework that extracts and utilizes event-specific information from both the event query and the provided video to refine MLLMs' response. We propose a unique mechanism that decomposes on-demand event queries into iconic actions. Subsequently, we employ models like CLIP and BLIP2 to predict specific timestamps for event occurrences. 
Our evaluation, conducted using the Charades-STA dataset, demonstrates a significant reduction in temporal hallucinations and an improvement in the quality of event-related responses. This research not only provides a new perspective in addressing a critical limitation of MLLMs but also contributes a quantitatively measurable method for evaluating MLLMs in the context of temporal-related questions.
\end{abstract}

\begin{keywords}
Multimodal language model, Temporal hallucination, Correction, Iconic Action 
\end{keywords}

\section{Introduction}
\label{sec:intro}

\begin{figure}[t]
\begin{minipage}[b]{1.0\linewidth}
  \centering
\centerline{\includegraphics[width=7.0cm]{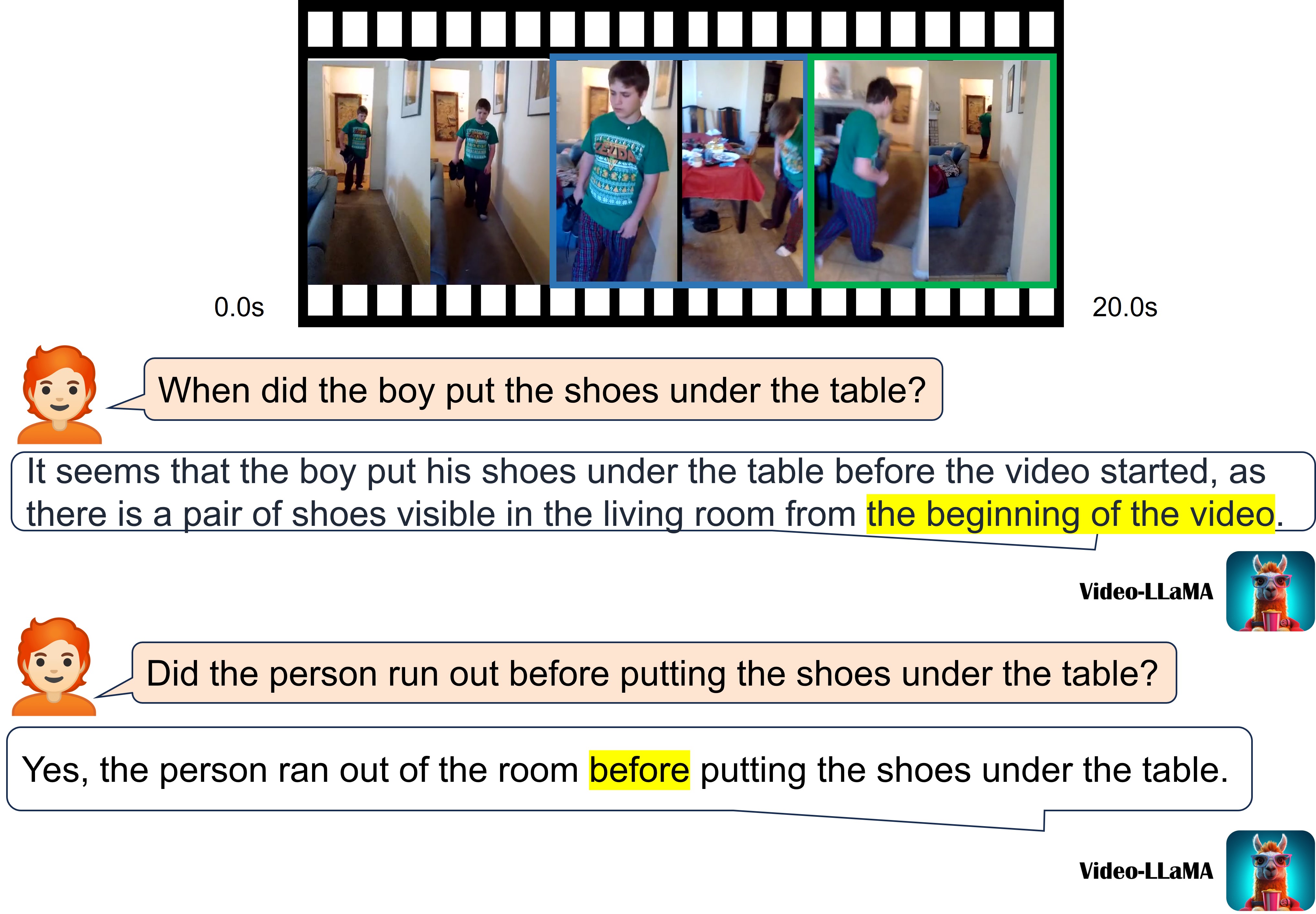}}
\end{minipage}
\caption{Examples illustrating hallucination generated by MLLMs in predicting event occurrence timestamps and sequencing.}
\label{fig:example1}
\end{figure}

Foundational large-scale models like CLIP~\cite{radford2021learning}, BLIP series~\cite{li2022blip, li2023blip}, and GPT series ~\cite{brown2020language, DBLP:conf/nips/Ouyang0JAWMZASR22} have ushered in a new era in multimodal understanding, transcending traditional classification tasks. Open-source large language models such as Vicuna~\cite{chiang2023vicuna} and LLaMA series~\cite{DBLP:journals/corr/abs-2302-13971, DBLP:journals/corr/abs-2307-09288} extended these capabilities, resulting in the development of Multimodal Large Language Models (MLLMs)~\cite{zhu2023minigpt, chen2023minigpt, liu2023visual, liu2023improved, zhang2023video} designed to enhance the comprehension of multimedia content.

Despite their proficiency in generating captivating descriptions, MLLMs often encounter challenges in vision tasks, leading to hallucinations ~\cite{rawte2023survey, zhang2023siren, yin2023woodpecker}—false perceptions of objects or events. While existing correction methods primarily target object-level hallucination~\cite{yin2023woodpecker, li2023evaluating}, There is no study focus on addressing event-level hallucination in MLLMs processing video inputs.

The temporal dimension introduced by videos sets them apart from image inputs, leading to temporal hallucination in MLLMs, especially in on-demand event queries. On-demand event queries refer to specific requests for information or data related to certain events or occurrences in videos, initiated at a user’s discretion.
To enhance the performance of MLLMs on diverse event queries presented by users, thoroughly analysis and understanding of  video content are crucial. However, this extensive amount of information could trigger a context-size limitation within the transformer mechanism, 
becoming notably significant
when processing large numbers of frames.

Figure  ~\ref{fig:example1} illustrates hallucinations in predicting event occurrence timestamps and temporal sequencing, generated by Video-LLaMA~\cite{zhang2023video} using raw video inputs.  
Constrained by the context-size limitation and training costs, Video-LLaMA uniformly samples videos at a fixed frequency for comprehension, inevitably resulting in some information loss.
The suboptimal performance of Video-LLaMA is caused by missing crucial on-demand event information in raw videos, exacerbated by lower sampling frequencies. 

To tackle the challenge of on-demand information loss in videos causing event-level hallucinations, we discuss and address three key questions related to MLLM's hallucinations at the event level: 1. What responses from MLLMs should be corrected? 2. How severe are these types of hallucinations? 3. What corrective measures should be employed to rectify them?

Regarding Question 1, our investigation highlights that
MLLMs exhibit subpar performance in accurately predicting the temporal location of on-demand information, consequently limiting their ability to predict the sequence of multiple events (which requires understanding the temporal location of each event). 
As a result, MLLMs are prone to temporal hallucinations in the mentioned scenarios.

We then have devised two tasks for a quantitative evaluation of MLLM's limitation in predicting the temporal location of events (Question 2). Task 1 involves predicting the precise timestamp of on-demand event occurrences, while Task 2 focuses on predicting the sequence of multiple events.

\begin{figure}[htp]
\begin{minipage}[b]{1.0\linewidth}
  \centering
\centerline{\includegraphics[width=1.0\linewidth]{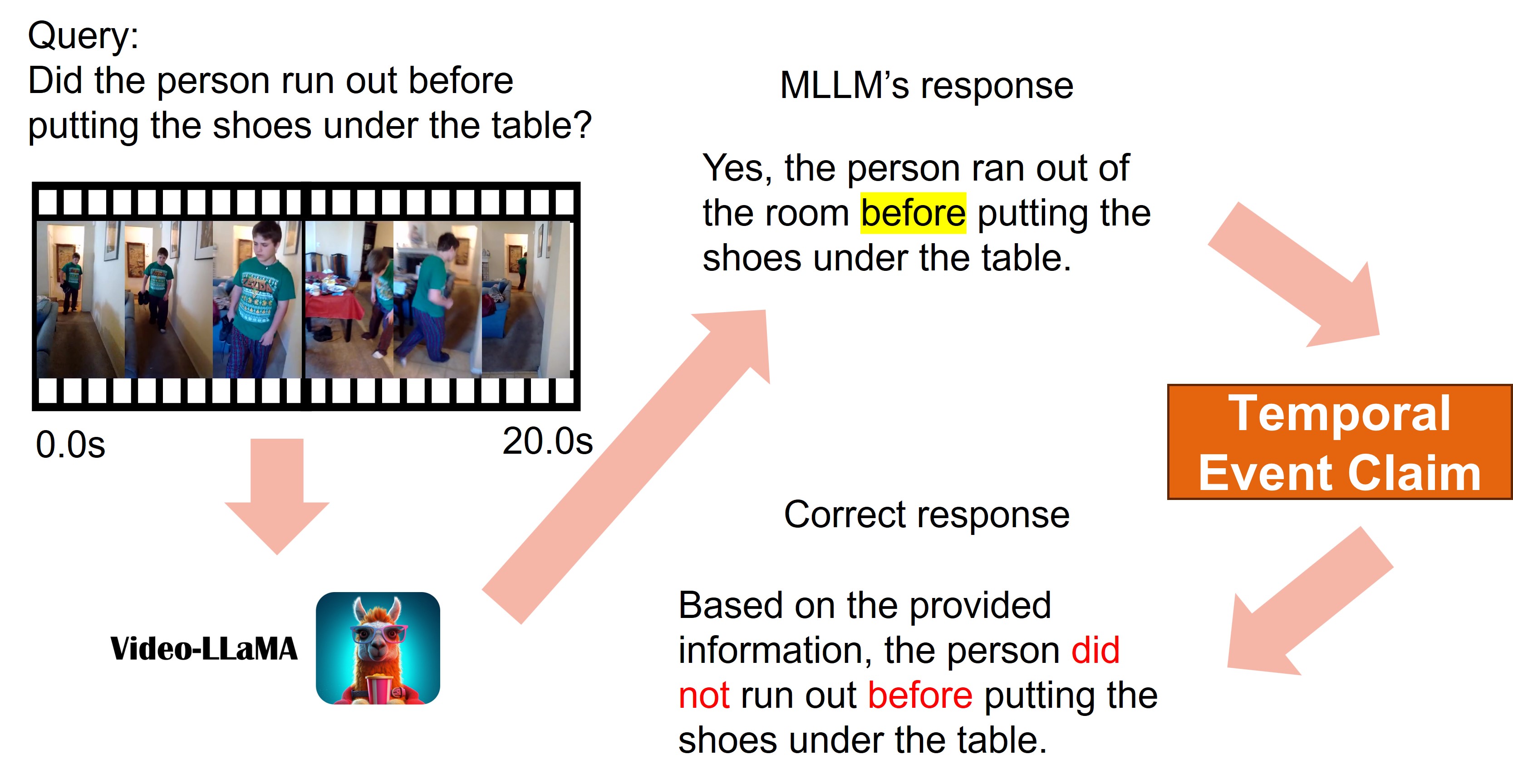}}
\end{minipage}
\caption{Framework overview of our temporal hallucination mitigating method.}
\label{fig:overview}
\end{figure}

To correct temporal hallucinations, we propose a novel method that incorporates corresponding event information, contributing to the generation of accurate temporal event claims (see Figure ~\ref{fig:overview} ). 
CLIP ~\cite{radford2021learning} and BLIP2 ~\cite{li2023blip} is used as external tools to extract specific temporal information from the frames that match the on-demand events.
Subsequently, we utilize the generated claim with event information to correct MLLMs' responses of questions related to event temporal information.
The time information from matched frames significantly reduces hallucinations in answering questions about event occurrence times and sequencing.

The study makes following contributions:

1. We introduce a novel framework to alleviate temporal hallucinations in MLLMs when addressing on-demand event temporal queries. 

2. We develop a quantitative evaluation method to assess the effectiveness of MLLMs in handling temporal-related questions, specifically those related to event occurrence times and sequencing. 

3. Our method is notable for being training-free, low-cost and interpretable. 

\section{Research background}

\subsection{MLLMs}
Multimodal Large Language Models (MLLMs) are advanced AI models that integrate and process information from multiple modalities, such as text, images, and videos, to perform a variety of tasks. Unlike traditional language models which primarily focus on text-based inputs and outputs, MLLMs are designed to understand and generate content that involves both language and other forms of data, such as visual or auditory inputs \cite{yin2023survey}.

Currently, most MLLMs that support visual modality only accept image inputs \cite{zhu2023minigpt, chen2023minigpt, liu2023visual, liu2023improved}. 
We choose Video-LLaMA \cite{zhang2023video} as our baseline model for it can supports images, videos and audios as input.

\subsection{Hallucination in MLLMs}

With the development of generative AI technology, the issue of hallucination has gradually gained attention. For MLLMs and LLMs, hallucination refers to the model erroneously perceiving its output as correct ~\cite{rawte2023survey, zhang2023siren, yin2023woodpecker}. In the context of MLLMs, hallucination can be categorized based on content into object-level and event-level, and based on the reasons for hallucination into knowledge-deficiency and inductive-bias types.

\textbf{Object-level hallucination.}
Object-level hallucination refers to a phenomenon in which machine learning models, particularly MLLMs and LLMs, generate incorrect or distorted outputs related to object recognition. In this context, hallucination occurs when the model mistakenly perceives or includes objects in its generated outputs that do not exist in the input data or misinterprets their characteristics. 

Liu et al. introduced an evaluation and correction method to address object-level hallucination ~\cite{li2023evaluating}. Additionally, Yin et al. proposed a train-free method specifically designed for object-level hallucination, as detailed in their work ~\cite{yin2023woodpecker}.

\textbf{Event-level hallucination.}
Currently, there is a gap in research regarding event-level hallucination. To the best of our knowledge, this paper is the first to specifically address event-level hallucination in MLLMs. Our work concentrates on events within videos, specifically examining hallucinations that arise when posing temporal-related queries. We have defined two tasks—event occurrence time and the order of occurrences for multiple events—to evaluate event-level temporal hallucination.

\subsection{Hallucination correction}

Dhuliawala et. al. divides hallucination correction methods into three categories: training-time correction, generation-time correction and correction based on external tools ~\cite{dhuliawala2023chain}.

While using external tools to address hallucination. The typical method is retrieval augmented generation (RAG) ~\cite{DBLP:conf/emnlp/0001PCKW21}, fact tool ~\cite{chern2023factool} and chain-of-thought verification ~\cite{zhao2023verify}.
 
Our approach is akin to RAG and fact tool, involving the extraction of on-demand event information via external tools to enhance the generation performance of MLLMs in addressing video event temporal-related questions.

\section{Mehthod}
\label{sec:method}
 
We employ Video-LLaMA ~\cite{zhang2023video} as a base MLLM, 
which is an MLLM extending the capabilities of LLMs to the processing and understanding of video content. It integrates two key branches in its architecture: a Vision-Language branch for dealing iwth input video frames, and an Audio-Language branch for handling input audio signals. This architecture enables Video-LLaMA to comprehend both visual and auditory elements in videos. 
It incorporates position embeddings 
in these inputs to encode their temporal information, enabling it to understand and interpret the sequence and timing of events in videos.

The rest of this section is organized as follows. The two tasks for temporal hallucination evaluation will be introduced in Sec. ~\ref{sec:hallucination}.
The general event temporal hallucination correction will be introduced in Sec. ~\ref{sec:correction}.
Response correction will be introduced in Sec. ~\ref{sec:response}. 

\subsection{Tow tasks for temporal hallucination evaluation}
\label{sec:hallucination}

As previous research has not extensively covered temporal hallucination for MLLMs, we construct test samples from available datasets and incorporate two tasks to evaluate an MLLM's vulnerability to temporal hallucination. Both of these tasks 
have the form of Video Question Answering (VQA) tasks.
The first task (Task 1) involves predicting the timestamp of event occurrences, with typical query questions such as \emph{When does/did the event ... occur?} The second task (Task 2) involves predicting the order of occurrences for multiple events, with typical questions like \emph{Did event A occur before/after event B?}  

\subsubsection{Timestamp prediction of event occurrences}

\begin{figure}[htp]
\begin{minipage}[b]{1.0\linewidth}
  \centering
\centerline{\includegraphics[width=1.0\linewidth]{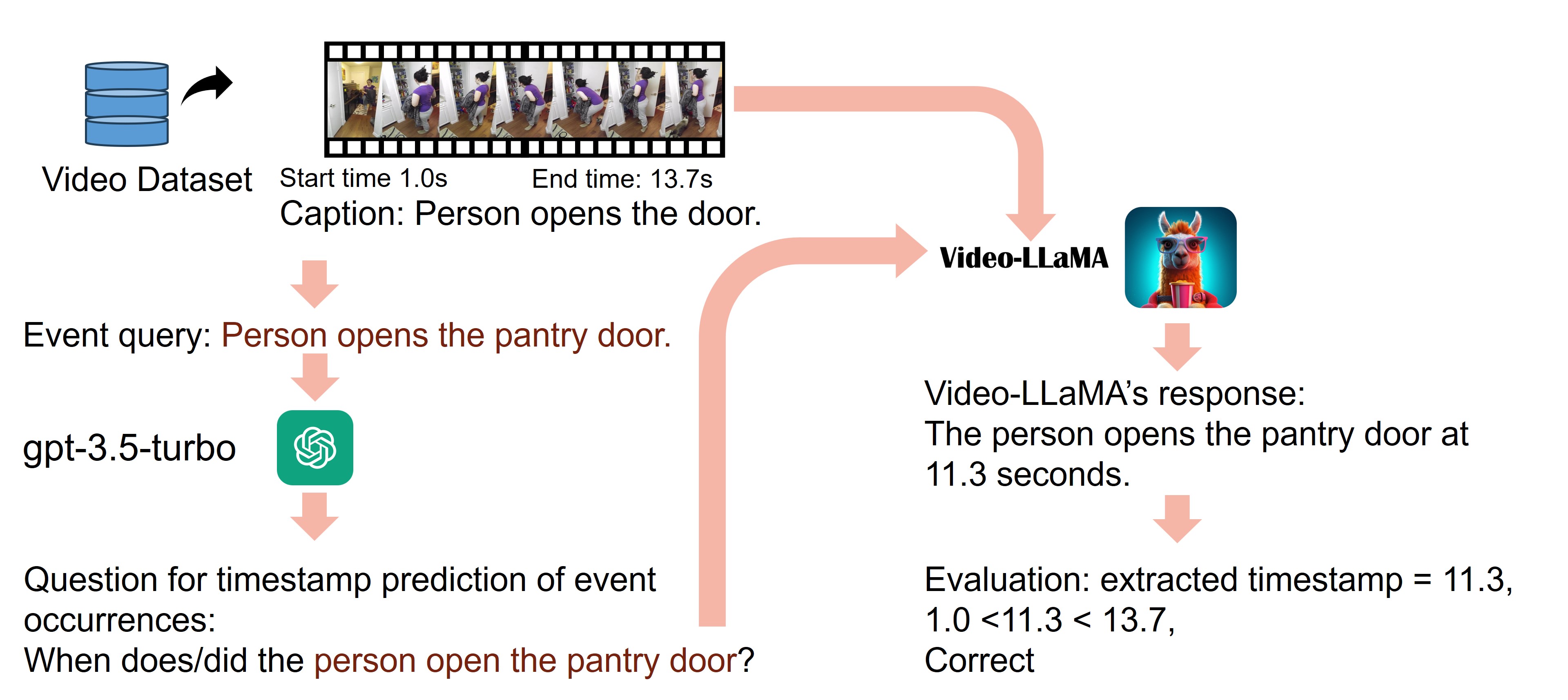}}
\end{minipage}
\caption{Evaluation for Task 1.}
\label{fig:task_1}
\end{figure}

An event query is defined as the caption of a specific video moment with known start and end timestamps. We source these event queries and their corresponding video moments from existing video datasets with temporal annotations.

The evaluation procedure of Task 1 is shown in Figure \ref{fig:task_1}.
For a given event query, we use GPT-3.5-turbo and a prompt (provided in Sec. 1.1 in Supplementary)
to transform the event query into a question regarding the appropriate occurrence of the event. Subsequently, we input this question into the MLLM, extract the response, and perform a thorough evaluation of the obtained answer.

In practice, the types of answers from MLLMs are diverse. They may provide one or more precise timestamps, a time duration, or even vague answers like \emph{at the beginning} or \emph{the end.} (see Table ~\ref{table:eval_example_task1}). To handle this diversity, we use GPT-3.5-turbo to parse these responses into a set of precise timestamps for evaluation (See prompt in Sec. 1.2 in Supplementary ). Some examples from the MLLM's responses and the transformed timestamp set are shown in Table ~\ref{table:eval_example_task1}. 
The set of timestamps $\{t\}$ will be evaluated accurately, without the need for subjective judgment of broad textual information.
 
\begin{table}[h]
\begin{center}
\footnotesize
\caption{Examples of response transformation for the evaluation of Task 1.} 
\label{table:eval_example_task1}
\begin{tabular}{|p{6cm}|p{2cm}|}
  \hline
  Response example &  Extracted timestamp set \\
  \hline
  Person opens the door at 3.2 second, 4.5 second. & $t \in $ [3.2, 4.5]  \\
  Person opens the door from 3.2 second to 4.5 second. & $3.2 \leq t \leq 4.5$  \\
  Person opens the door in the beginning of the video & $0.0 \leq t \leq \frac{1}{3}L$  \\
  Person opens the door in the middle of the video & $\frac{1}{3}L \leq t \leq \frac{2}{3}L$  \\
  Person opens the door in the end of the video & $\frac{2}{3}L \leq t \leq L$  \\
  Person opens the door throughout the video & $0.0 \leq t \leq L$  \\
  No information mentioned. & $t \in \emptyset$ \\
  \hline
\end{tabular}
\end{center}
\end{table}

\subsubsection{Order Prediction of Event Occurrences}

\begin{figure}
\begin{minipage}[b]{1.0\linewidth}
  \centering  \centerline{\includegraphics[width=8.0cm]{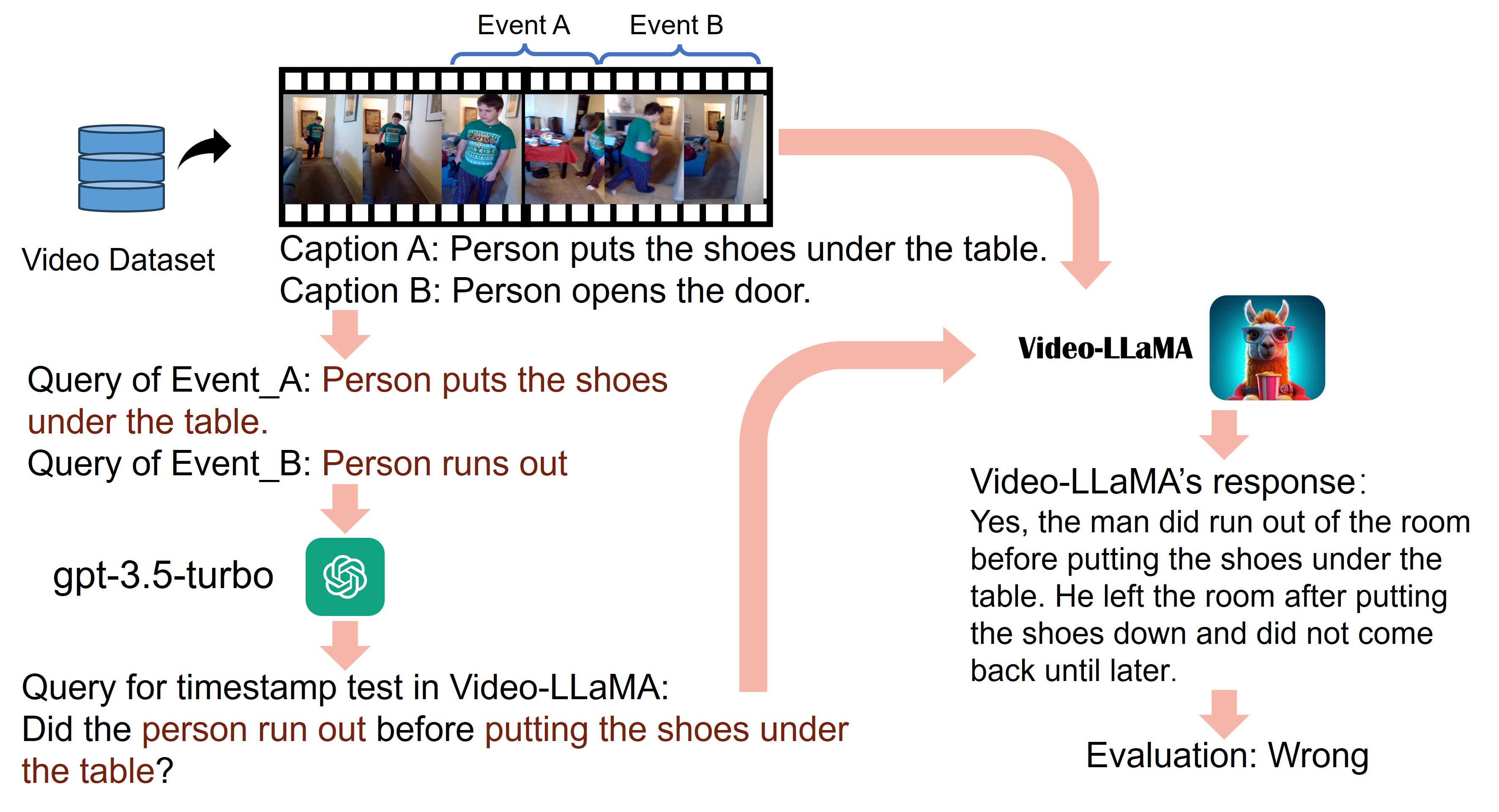}}
\end{minipage}
\caption{Evaluation for Task 2.}
\label{fig:task2}
\end{figure}

Task 2 is defined to evaluate the performance on predicting orders of event occurrences (see Figure \ref{fig:task2}).
The queries for Event A and B are derived from two temporally different captions, each with its own set of start and end time annotations. To evaluate each video, we randomly select two events, shuffle their order, and then utilize GPT-3.5-turbo to formulate questions like \emph{Did event A occur before/after event B?}. Subsequently, we engage MLLMs to respond to the questions we generate, and assess the answers based on the time annotations of the two events.
 
The prompt used for generating the event order questions is provided in Sec. 1.3 in Supplementary.

For this type of question, the answers generated by the original MLLM can be categorized into three classes: \textbf{Yes}, \textbf{No}, and \textbf{No relevant information}. 

Comparing the ground-truth temporal locations of the two events, the correct answers for the event orders can only be either \textbf{Yes} or \textbf{No}.

\subsection{Event temporal
hallucination correction}
\label{sec:correction}

\begin{figure}[t]
\begin{minipage}[b]{1.0\linewidth}
  \centering  \centerline{\includegraphics[width=1.0\linewidth]{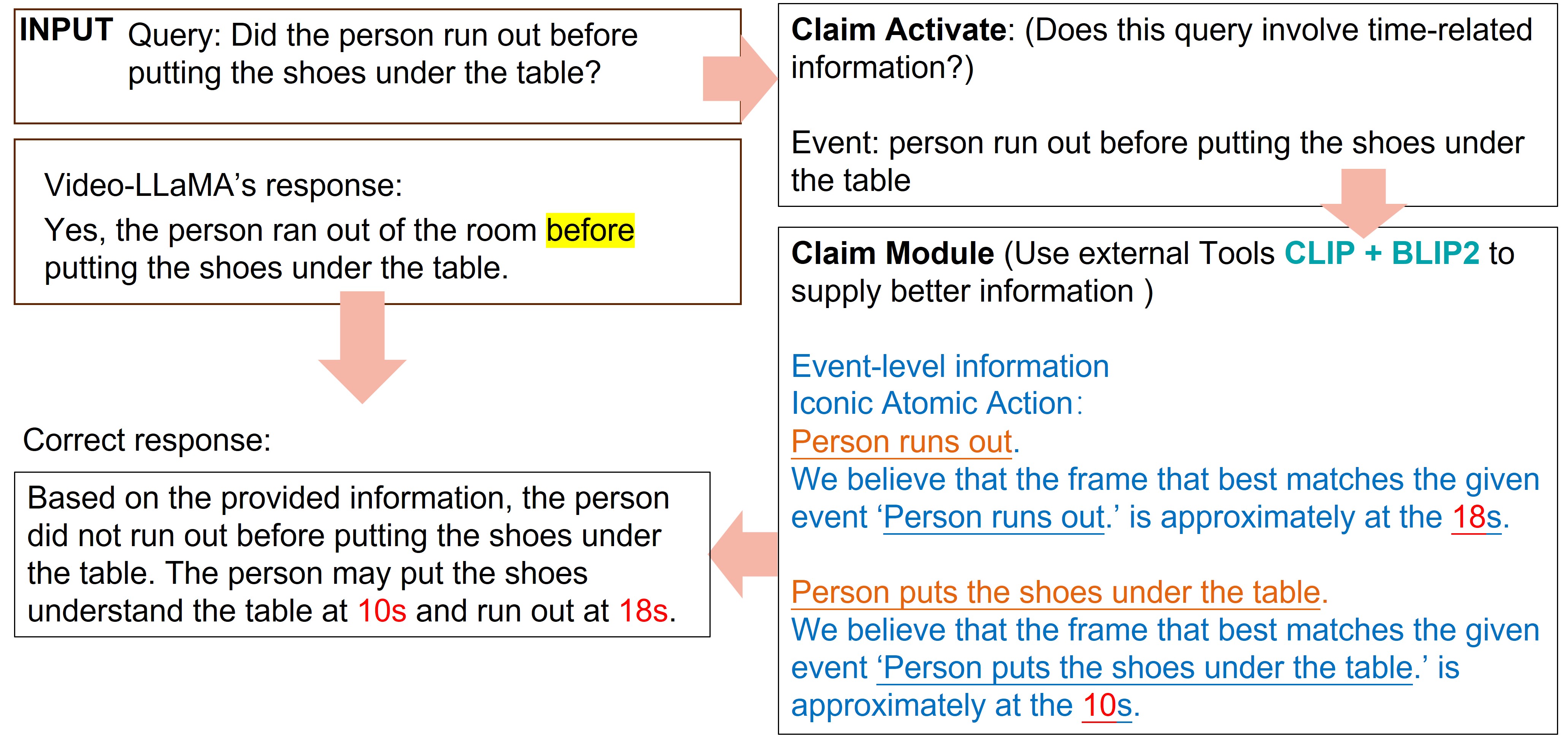}}
\end{minipage}
\caption{Illustration of event temporal hallucination correction.}
\label{fig:Claim}
\end{figure}

For each instance of temporal hallucination, we need to generate a claim. This claim acts as a standardized template for inputting corrective information, which is used to revise the MLLM’s response.
It consists of two components: Claim Activate and Claim Module. 
Figure ~\ref{fig:Claim} illustrates the correction process involving two essential components. 
In the initial correction step, the Claim Activate component takes the user's query as input and utilizes GPT-3.5-turbo to determine if the query requires temporal information support. Additionally, it detects the events in the query, and the identified event text serves as input for the Claim Module.

The Claim Module generates a Claim for correcting temporal hallucinations based on the inputted events. We have designed an external tool using CLIP and BLIP2 to obtain specific event temporal information. After filling in the template with this information, the Claim is generated in the Claim Module. This approach ensures that the correction process is informed by accurate and on-demand temporal details, mitigating the temporal hallucination in the MLLM's responses.

In Figure ~\ref{fig:Claim}, the text enclosed by dotted lines represents the claim template, and the portions with underlines indicate the corresponding event information to be filled in based on external tools.

The whole claim module can be organized as two steps:

1. Decompose the given event description to several iconic actions. This step is to improve the event prediction precision via CLIP-like external tools. We thus decompose the original event query to multiple ``Iconic Actions'', which refer to actions with visual representations that are easily recognized by image-based vision language models such as CLIP.

2. Provide the frame when the iconic event most likely occurred. In this step we find the frame when the iconic event most likely occurred lerverage CLIP and BLIP2. Then we can predict the timestamp of the frame as the specific occurrance time of the give event.
The timestamps we predict will be utilized as factual evidence to populate the claim template.

\subsubsection{Decompose event to iconic actions}
To improve the event prediction precision through CLIP-like external tools, 
the provided event description is decomposed into distinct ``Iconic Actions''.
``Iconic Action'' refers to distinctive, easily recognizable actions or events within a video that are emblematic of the content or narrative, aiding in quick comprehension and contextual understanding for viewers. 
For example, in a video of a football match, iconic actions might be captioned as follows: \emph{The player kicks the ball, it sails past the goalkeeper, and lands in the net, followed by the crowd's loud cheers.} 
This description involves four iconic actions: 
\emph{
1. The player kicks the ball. 
2. The ball sails past the goalkeeper. 
3. The ball lands in the net.
4. The crowd's loud cheers.}

With human common sense, we can easily envision stereotypical images of these four iconic actions. It is also easy to visually separate each action by observing dynamic changes such as the position of the player's foot, the location of the ball, and the state of the audience.
GPT-3.5-turbo is used to discern and isolate key components, laying the groundwork for a more nuanced understanding of the event. 
The prompt is listed in in Sec. 2.1 in Supplementary. 

\subsubsection{Timestamp identification for iconic actions}
\label{sec:timestamp}
\begin{figure}[t]
\begin{minipage}[b]{1.0\linewidth}
  \centering \centerline{\includegraphics[width=1.0\linewidth]{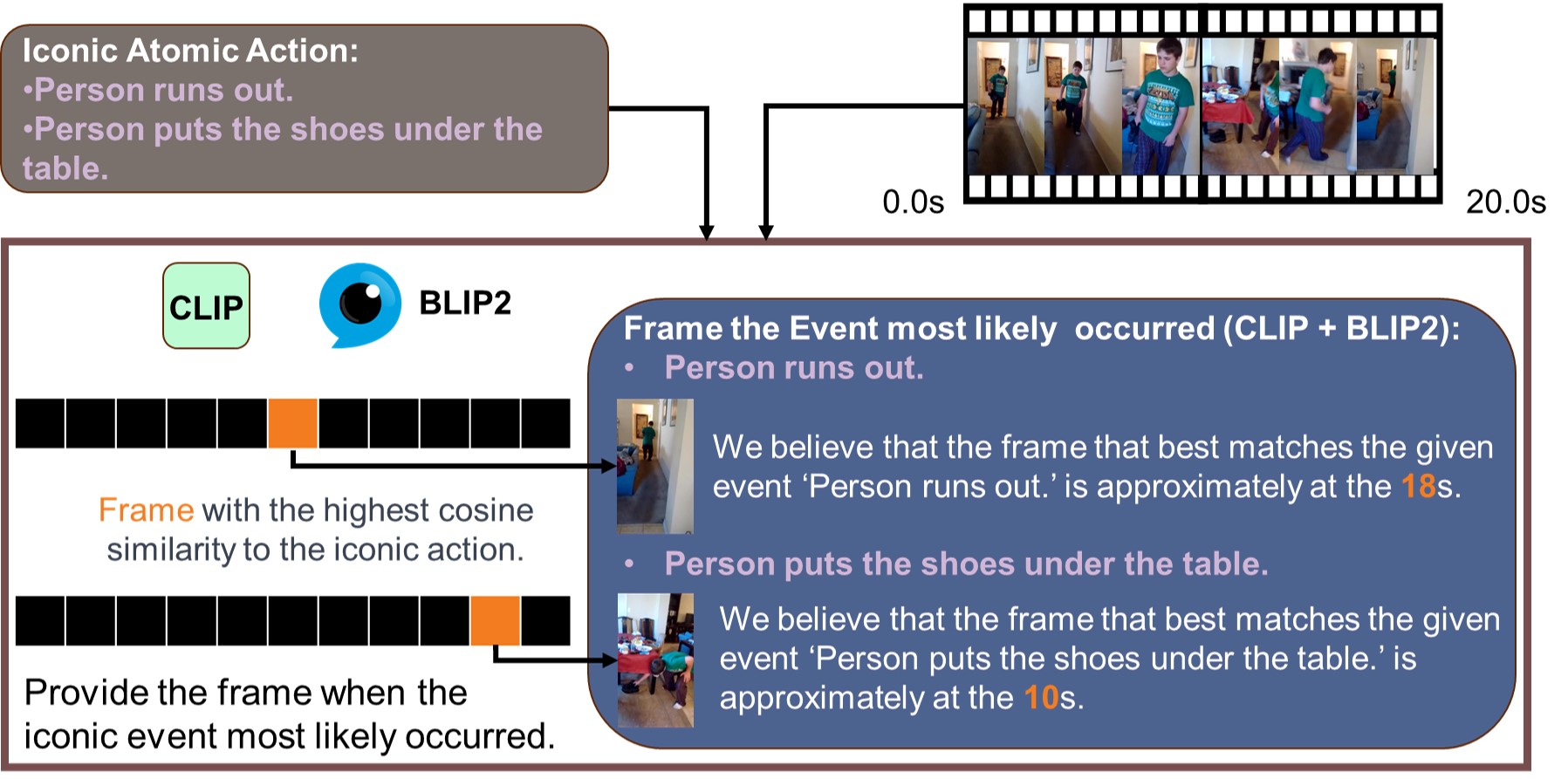}}
\end{minipage}
\caption{Illustration of timestamp identification for iconic actions.}
\label{fig:frame_pred}
\end{figure}

To identify the timestamps of iconic actions, we provide the frame when the iconic action most likely occurred. Figure ~\ref{fig:frame_pred} illustrates the produce of timestamp identification for iconic actions.

We utilize CLIP and BLIP2 to calculate the matching scores between each frame and the iconic event text. Subsequently, we choose the frame with the highest score.
Denoting the video has $N$ frames and the $k^{th}$ frame is $I_k$, the $j^{th}$ iconic action of the on-demand query is $Q_j$. 
The CLIP's image and text match score can be denoted as $\mathrm{cos_{\text{CLIP}}}$ and BLIP2's  is $\mathrm{cos_{\text{BLIP2}}}$. 

\begin{equation}
    T_{\tau} = \mathop{\mathrm{argmax}}_{k=1, ..., N}
    (\mathrm{cos_{\text{CLIP}}}(I_k, Q_j) +\mathrm{cos_{\text{BLIP2}}}(I_k, Q_j)),
\label{eq:cos1}   
\end{equation}
where $T_{\tau}$ is the most representative timestamp for $j^{th}$ iconic action.

To further improve frame matching performance, we employ the test-time distribution normalization method \cite{zhou2023test} to enhance CLIP's matching performance. We normalize the image and text feature of the current $Q_j$ and all frames ${I_k}, k=1,2, ...,N$. The new CLIP's matching score can be computed as:

\begin{equation}
S_{DN} = \cos_{\text{CLIP}}(I_k - \lambda \mu_{I}, Q_j - \lambda \mu_{Q_j}),
\label{eq:cos2}   
\end{equation}
where $\mu$ is the mean value. $\lambda$ is set as 0.25, the same as \cite{zhou2023test}.

Finally, the we have matching score like:
\begin{equation}
\begin{aligned}
     T_{{\tau}^\ast} = \mathop{\mathrm{argmax}}_{k=1, ..., N}
    (\mathrm{cos_{\text{CLIP}}}(I_k, Q_j) \\
    +\mathrm{cos_{\text{BLIP2}}}(I_k, Q_j) 
    + S_{DN}),
\end{aligned}
\label{eq:cos3}  
\end{equation}
where $T_{{\tau}^\ast}$ is the predict timestamp for the give iconic action $Q_j$. Both $Q_j$ and $T_{{\tau}^\ast}$ will be filled into the claim template.

\subsection{Response correction}
\label{sec:response}

\begin{figure}[t]
\begin{minipage}[b]{1.0\linewidth}
  \centering \centerline{\includegraphics[width=1.0\linewidth]{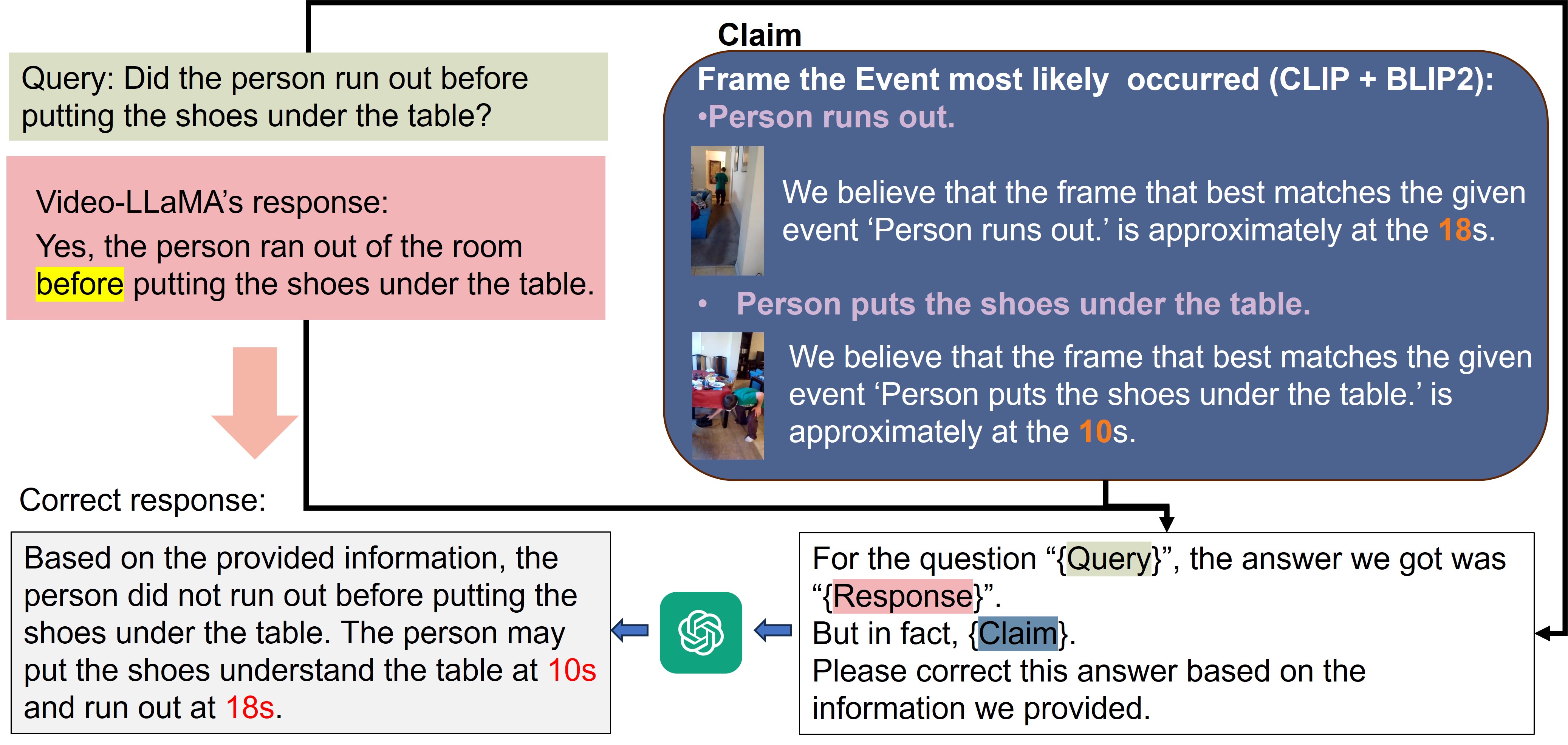}}
\end{minipage}
\caption{Illustration of response correction.}
\label{fig:response}
\end{figure}

The user's query, MLLM's response, and the generated claim are utilized to derive the new corrected response. Figure ~\ref{fig:response} illustrates the procedure of correcting the MLLM's response using the claim generated from steps 1 and 2 in the claim module.
A corrective prompt (see Figure ~\ref{fig:response}), comprising the user's query, MLLM's response, the generated claim, and GPT-3.5-turbo, is utilized to generate the updated response.

\section{Experiment}

\subsection{Experiment setting}
\textbf{Dataset.}
The evaluation can utilize open video datasets containing captions and corresponding temporal annotations.
Charades-STA~\cite{gao2017tall} is a widely used temporal activity dataset in the field of moment retrieval and temporal sentense grounding.
In Section ~\ref{sec:hallucination}, we offer detailed explanations of how we use videos with captions and temporal annotations to create our specific evaluation tasks.
For the Timestamp Prediction task, we evaluate using all sentences in the test set (3,720 sentences from 1,334 videos) of Charades-STA.
For the Order Prediction task,
We initially filter 637 videos and then randomly generate 1272 questions containing \emph{before} or \emph{after}. The temporal overlap between each pair of events in these questions is less than 0.5.

\textbf{Baseline model.}
Video-LLaMA \cite{zhang2023video} with Llama-2-7B-Chat as language decoder
is used as the baseline MLLM for it supports video and image input. 
GPT-3.5-turbo is used as LLM.
CLIP ViT-L/14 336px \cite{radford2021learning} and BLIP2 \cite{li2022blip} are used as external tools to get relevant frames' timestamps.

\textbf{Implementation details.}
Video-LLaMA's number of beam search is set as 1. The temperature is set as 0.1 and 1.0 for different experiments.
We use 1 FPS for extracting frames from video.

\textbf{Evaluation metrics.} 
In Task 1, a relaxed metric is used to evaluate MLLM's various responses.
If the extracted timestamps set from the MLLM's response is represented as $\{t\}$, and the annotated start and end timestamps of the event are denoted as $T_s$ and $T_e$ respectively, the response is deemed correct if the condition $\exists t \in \{t\} : T_s \leq t \leq T_e$ is satisfied; otherwise, it is considered incorrect.
We use \textbf{R@1} and \textbf{R@5} as the evaluation metric of random experiment and corrected responses. \textbf{R@1} means only the frame timestamp with largest score in Equation ~\ref{eq:cos3} will be used for evaluation while \textbf{R@5} means the top 5 frame timestamps will be used for evaluation.
In Task 2, the MLLM's response is considered correct only if its categorized results match the ground truth class (\textbf{Yes} / \textbf{No}).

\subsection{Temporal hallucination evaluation and correction result}
We evaluate and correct the temporal hallucination on Task 1 and 2.

\textbf{Results on timestamp prediction of event occurrences.} 
The results for timestamp prediction of event occurrences are presented in Table ~\ref{table:timestamp}. Results marked with an asterisk in the Video-LLaMA column were obtained without restricting the output format of Video-LLaMA. 

Even with the relaxed evaluation criteria for Video-LLaMA, it can be observed that Video-LLaMA only marginally outperforms random temporal predictions. In contrast, our method significantly outperforms both random predictions and Video-LLaMA.

\begin{table}[htp]
\begin{center}
\footnotesize
\caption{Results on timestamp prediction} \label{table:timestamp}
\begin{tabular}{|c|c|c|}
  \hline
  Method & R@1 Acc & R@5 Acc \\
  \hline
  Random & 25.59 & 52.63 \\
  \hline
  Video-LLaMA \cite{zhang2023video} ($\mathrm{temp}=0.1$)& 
  \multicolumn{2}{c|}{ 29.57}\\
  \hline
  Video-LLaMA \cite{zhang2023video} 
  ($\mathrm{temp}=1.0$)& 
  \multicolumn{2}{c|}{ 29.81}\\
  \hline
  Hallucination-reduced MLLM (ours) & 57.66 & 85.29 \\
  \hline
\end{tabular}
\end{center}
\end{table}

\textbf{Results on order prediction of event occurrences.}  The results are shown in Table ~\ref{table:order}.
In terms of answer range, predicting the sequence of events is relatively easier. As shown in the table, the results indicate a clear improvement with our method compared to both random predictions and Video-LLaMA.

\begin{table}[htp]
\footnotesize
\begin{center}
\caption{Results on order prediction} 
\label{table:order}
\begin{tabular}{|c|c|c|}
  \hline
  Method &  Acc \\
  \hline
  Random & 24.20  \\
  \hline
  Original Video-LLaMA \cite{zhang2023video} & 49.21 \\
  \hline
  Hallucination-reduced MLLM (ours) & 67.53 \\
  \hline
\end{tabular}
\end{center}
\end{table}

\subsection{Ablation experiment for external tools}

We compared the performances of different tools—CLIP, BLIP2, and CLIP with subtracted mean values for both image and text—in determining timestamps (see Table ~\ref{tab:ablation1}). 
The experimental results indicate that ensemble these models can effectively enhance timestamp prediction performance.

\begin{table}[htb]
\footnotesize
  \centering
  \caption{Ensemble Tool Results Comparison }
  \label{tab:ablation1}
  \begin{tabular}{|l|c|c|c|}
    \hline
    \textbf{Model} & \textbf{R@1 Acc} & \textbf{R@5 Acc} \\
     \hline
    Original CLIP & 56.59 & 84.89 \\
     \hline
    BLIP2 & 54.67 & 83.49 \\
     \hline
    CLIPwithDN & 56.56 & 85.00 \\
     \hline
    CLIP + BLIP2 + CLIPwithDN & 57.66 & 85.29 \\
     \hline
  \end{tabular}
\end{table}

\section{Conclusion}

In this study, we have addressed a significant challenge in the realm of Multimodal Large Language Models (MLLMs) – the occurrence of event-level hallucinations, particularly when processing video inputs. 

By decomposing on-demand event queries into iconic actions and employing models like CLIP and BLIP2 for frame identification, our method has demonstrated a marked improvement in pinpointing event occurrences and understanding temporal sequences. 

Our evaluation with the Charades-STA dataset has shown a significant reduction in temporal hallucinations, thereby enhancing the accuracy and reliability of MLLMs in handling video content. The qualitative improvements observed in our study not only validate our approach but also pave the way for future research in this field.

\bibliographystyle{IEEEbib}
\bibliography{icme2023template}
\end{document}